\documentclass[fleqn,10pt]{wlscirep}
\usepackage{subcaption}
\usepackage{adjustbox} 
\usepackage{tabularx} 
\usepackage{multirow} 
\usepackage{siunitx}
\usepackage{makecell}
\usepackage{nicematrix}
\usepackage{authblk}

\usepackage[utf8]{inputenc}
\usepackage[T1]{fontenc}
\title{A cautionary tale on the cost-effectiveness of collaborative AI in real-world medical applications}

\author[1,+]{{Francesco} {Cremonesi}}
\author[1,2,+]{{Lucia} {Innocenti}}
\author[2]{{Vicky} {Goh}}
\author[2]{{Sebastien} {Ourselin}}
\author[2,=]{{Michela} {Antonelli}}
\author[1,=,*]{{Marco} {Lorenzi}}

\affil[1]{{Epione Research Group}, {Inria Center of Universit\'e C\^ote d'Azur}, {France}}

\affil[2]{{King’s College London}, {{London}, {UK}}}

\affil[+]{joint first authors}
\affil[=]{joint last authors}
\affil[*]{marco.lorenzi@inria.fr}

\begin{abstract}
Federated learning (FL) has gained wide popularity as a collaborative learning paradigm allowing collaborative Artificial Intelligence (AI) in sensitive healthcare applications. Nevertheless, the practical implementation of FL presents technical and organizational challenges, as it generally requires complex communication infrastructures. In this context, consensus-based learning (CBL) may represent a promising alternative for collaborative learning, allowing the combination of local knowledge into a federated decision system, while potentially reducing deployment overhead. Nevertheless, a comprehensive assessment of the viability of consensus-based learning as a cost-effective and sustainable alternative to federated learning has not yet been conducted.
In this work we propose an extensive benchmark of the accuracy, cost-effectiveness, and sustainability of a panel of FL and CBL methods in a wide range of collaborative medical data analysis scenarios. The benchmark includes 7 different medical datasets, encompassing 3 machine learning tasks, 8 different data modalities, and multi-centric settings involving 3 to 23 clients.
Our results reveal that CBL is a cost-effective and more sustainable alternative to FL. When compared across the panel of medical datasets in the considered benchmark, CBL methods provide equivalent accuracy to the one achieved by FL. Nonetheless, CBL significantly reduces training time and communication cost (resp. 15 fold and 131 fold decrease) ($p$-value < 0.05).
This study opens a novel perspective on the deployment of collaborative AI in real-world applications, whereas the adoption of cost-effective methods is instrumental to achieve sustainability and democratization of AI by alleviating the need for extensive computational resources.

\end{abstract}
\begin{document}

\flushbottom
\maketitle
%
%
\thispagestyle{empty}

\section*{Introduction}

Collaborative learning (CL) has become a popular paradigm for the exploitation of Artificial Intelligence (AI) in healthcare applications~\cite{sheller2020federated,rieke2020future}. CL allows multiple parties to jointly solve analytical tasks by combining local knowledge extracted from the respective data. These abstractions can, for instance, take the form of global data statistics, machine learning (ML) model parameters, or model’s predictions. This paradigm is rapidly gaining popularity in healthcare applications due to its appealing promises for accuracy/privacy trade-off~\cite{nature2021coll,usynin2021adversarial}.
Federated Learning (FL) has been identified as a key CL paradigm ~\cite{rieke2020future}, focusing on collaboratively optimizing model parameters across clients, each holding local datasets (Figure 1, left). FL is based on an iterative optimization paradigm in which each client shares model parameters partially trained on the respective local data, which are then aggregated by a central server to obtain a global model. Although FL has demonstrated significant advantages in security and privacy over centralized approaches, its implementation in real-world applications is not straightforward and can face technical and organizational challenges~\cite{kairouz2021advances}, reflecting the technology’s complexity as well as significant computational costs~\cite{joshi2022federated}.
Consensus-based learning (CBL)~\cite{guha2019one,wang2020federated,innocenti2023benchmarking} represents a relevant CL alternative to FL. CBL does not rely on a shared training routine nor on a common model architecture across parties, but instead combines the predictions obtained from the different models independently trained by each client on their local data. CBL thus relies on an off-line setting, in which information is exchanged only at inference time (Figure 1, right).
Bearing in mind the duality between these paradigms, when it comes to implementing collaborative frameworks, we currently lack quantitative benchmarks illustrating the quality and cost-effectiveness of specific CL settings in real-world medical applications. On the one hand, FL has been widely investigated for CL applications, in particular for healthcare~\cite{lee2020federated,chen2022pfl,caldas2018leaf}. On the other hand, while there exists a large body of literature demonstrating the effectiveness of CBL over training local models, its applications to distributed datasets remain under-investigated except for some recent works~\cite{guha2019one,chaudhari2023safenet,gupta2023collaborative}. To our knowledge, a general reference point that allows comparison between CBL and FL paradigms while considering the accuracy and cost-effectiveness of these paradigms is still missing.
Beyond the accuracy and cost of individual deployments, the choice of collaborative learning paradigm also carries broader implications for the sustainability of AI in healthcare, encompassing ecological considerations, such as the energy consumption associated with iterative model training and communication, as well as economic considerations related to the infrastructure and maintenance costs borne by participating institutions, which are especially relevant for healthcare providers operating under limited resources.
To fill this gap, in this study we assess the viability of consensus-based learning as compared to federated learning in healthcare applications, providing the first comparative analysis of the capabilities and cost-effectiveness of CBL and FL in an extensive  benchmark including a large panel of medical datasets, machine learning tasks and federated clients. 
Our results show that CBL achieves accuracy on par with FL while reducing training time and communication cost by up to 15-fold and 131-fold respectively, underscoring its potential as a more sustainable and cost-effective alternative for collaborative AI in healthcare.

The main contributions of this work can be summarized as follows: (i) we provide the first systematic benchmark comparing CBL and FL across a broad panel of medical datasets, spanning diverse machine learning tasks, data modalities, and numbers of participating clients; (ii) we quantitatively demonstrate that CBL achieves accuracy comparable to FL while substantially reducing training time and bandwidth usage; and (iii) we discuss the implications of these findings for the sustainable and cost-effective deployment of collaborative AI in real-world healthcare settings.

\section*{Results}

We identified a set of seven dataset-task pairs that constitute a representative benchmark of multi-centric collaborations in medical applications. 
The benchmarks present a high degree of heterogeneity in terms of the number of clients in the federation (from 3 to 23), sample sizes (100 to over 2000 patients), data modalities (total of 8 different modalities), and distribution of data among clients, thus reflecting the heterogeneity of real-world applications. 
The main features of each task are summarized in Table~\ref{tab:dataset}.

\begin{table}
  \centering
  \caption{Datasets used for the benchmark. For each dataset, we report the number,  the size of clients' local datasets, the data modality, and the task, along with the associated metric.}
  \label{tab:dataset}
  \begin{tabularx}{\linewidth}{@{}ccccc@{}}
    \toprule
    \multirow{3}{*}{\parbox{2cm}{\centering Dataset }} & \multirow{3}{*}{\parbox{1.7cm}{\centering \#Clients}} & \multirow{3}{*}{\parbox{3cm}{\centering Dataset size \\ per client}} & \multirow{3}{*}{Data Modality} & \multirow{3}{*}{\parbox{3cm}{\centering Task \\ \textit{[Metric]}}} \\
    & & & & \\
    & & & & \\
    \midrule
    \multirow{2}{*}{FedProstate \cite{antonelli2022medical,litjens2014evaluation,armato2018prostatex}} & \parbox{0.6cm}{\centering 6} & \multirow{2}{*}{\parbox{3cm}{\centering [32, 23, 27,\\ 184, 5, 36]}} & T2 MRI & \multirow{2}{*}{\parbox{3cm}{\centering Segmentation\\\textit{[DiceScore]}}} \\
    & & & & \\
    \midrule
    \multirow{2}{*}{FedHeart \cite{terrail2022flamby}} & \parbox{0.6cm}{\centering 4} & \multirow{2}{*}{\parbox{3cm}{\centering [303, 261,\\ 46, 130]}} & Tabular Data & \multirow{2}{*}{\parbox{2cm}{\centering Classification\\\textit{[Accuracy]}}} \\
    & & & & \\
    \midrule
    \multirow{2}{*}{FedIXI \cite{terrail2022flamby}} & \parbox{0.6cm}{\centering 3} & \multirow{2}{*}{\parbox{3cm}{\centering [311, 181, 74]}} & T1 MRI & \multirow{2}{*}{\parbox{2cm}{\centering Segmentation\\\textit{[DiceScore]}}} \\
    & & & & \\   
    \midrule
    \multirow{2}{*}{FedISIC \cite{terrail2022flamby}} & \parbox{0.6cm}{\centering 6} & \multirow{2}{*}{\parbox{3cm}{\centering [12k, 3.9k, 3.3k,\\225, 819, 439]}} & \multirow{2}{*}{\parbox{1.5cm}{\centering Dermoscopy\\ \centering images}} & \multirow{2}{*}{\parbox{2cm}{\centering Classification\\\textit{[Accuracy]}}} \\
    & & & & \\    
    \midrule
    \multirow{2}{*}{FedTCGA-BRCA \cite{terrail2022flamby}} & \parbox{0.6cm}{\centering 6} & \multirow{2}{*}{\parbox{3cm}{\centering [311, 196, 206,\\ 162, 162, 51]}} & Tabular Data & \multirow{2}{*}{\parbox{2cm}{\centering Survival\\\textit{[C-Index]}}} \\
    & & & & \\   
    \midrule
    \multirow{2}{*}{FedKiTS \cite{terrail2022flamby}} & \parbox{0.6cm}{\centering 6} & \multirow{2}{*}{\parbox{3cm}{\centering [12, 14, 12,\\ 12, 16, 30]}} & CT Scans & \multirow{2}{*}{\parbox{2cm}{\centering Segmentation\\\textit{[DiceScore]}}} \\
    & & & & \\   
    \midrule
    \multirow{2}{*}{FeTS \cite{pati2022federated}} & \parbox{0.6cm}{\centering 23} & \multirow{2}{*}{\parbox{3.2cm}{\centering $4$ to $511$ \\ \textit{
    (see Suppl.)}}} & \multirow{2}{*}{T1, T1CE, T2 \& FLAIR MRIs} & \multirow{2}{*}{\parbox{2cm}{\centering Segmentation\\\textit{[DiceScore]}}} \\
    & & & & \\       
    \bottomrule
  \end{tabularx}
\end{table}

\begin{figure*}[ht]
  \centering

  \begin{subfigure}{0.3\textwidth}
    \centering
    \includegraphics[width=\linewidth]{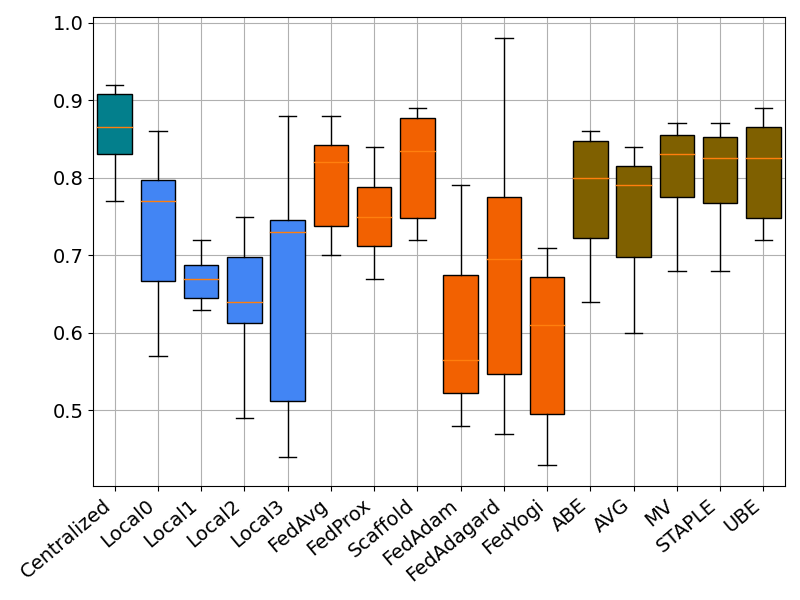}
    \caption{FedProstate}
    \label{fig:ac_prostate}
  \end{subfigure}
  \hfill
  \begin{subfigure}{0.3\textwidth}
    \centering
    \includegraphics[width=\linewidth]{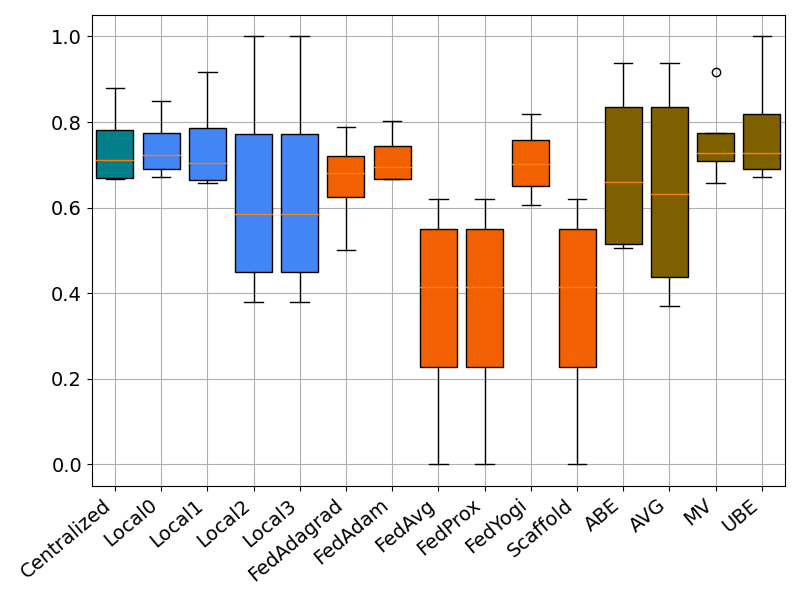}
    \caption{FedHeart}
    \label{fig:acc_heart}
  \end{subfigure}
  \hfill
  \begin{subfigure}{0.3\textwidth}
    \centering
    \includegraphics[width=\linewidth]{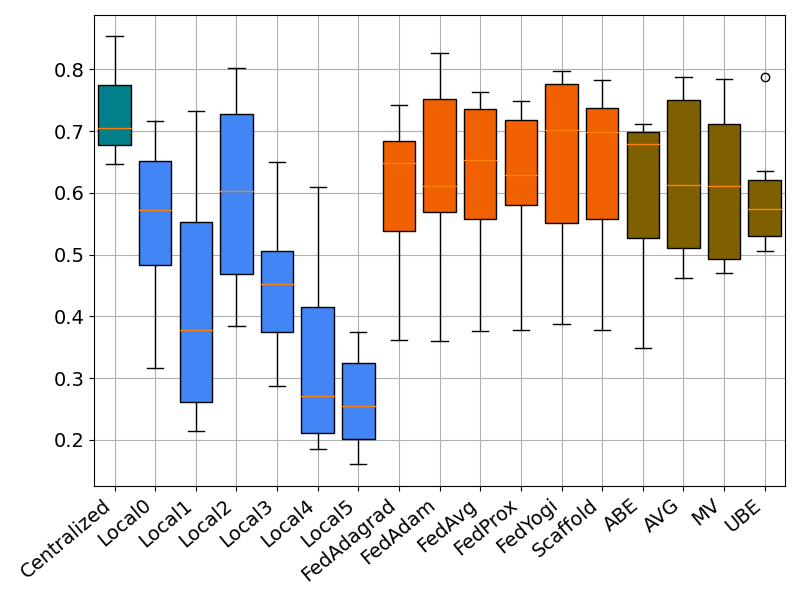}
    \caption{FedIsic}
    \label{fig:acc_isic}
  \end{subfigure}

  \vspace{12pt}

  \begin{subfigure}{0.3\textwidth}
    \centering
    \includegraphics[width=\linewidth]{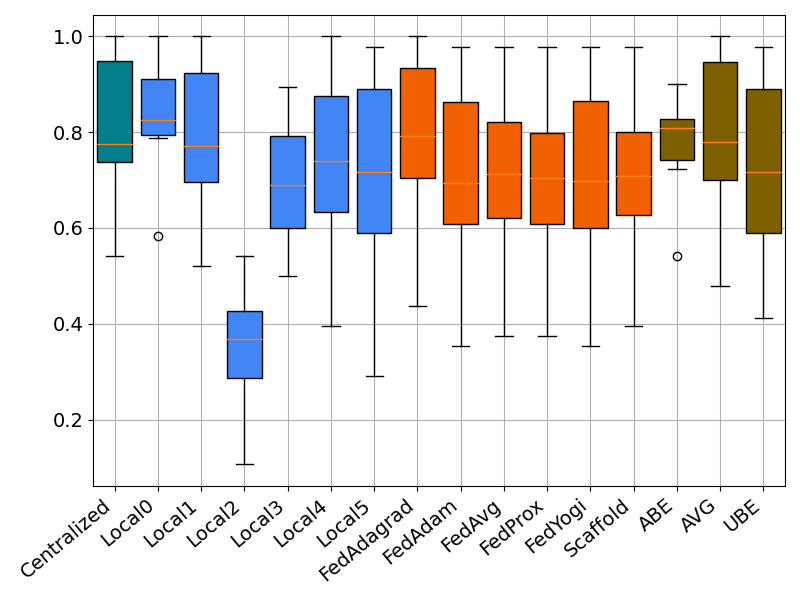}
    \caption{FedTGCA-BRCA}
    \label{fig:acc_tgca}
  \end{subfigure}
  \hfill
  \begin{subfigure}{0.3\textwidth}
    \centering
    \includegraphics[width=\linewidth]{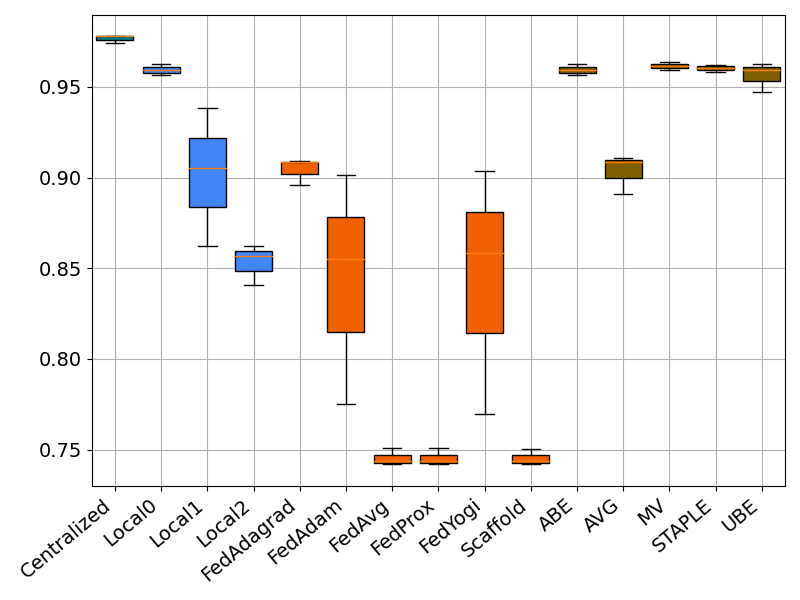}
    \caption{FedIXI}
    \label{fig:acc_ixi}
  \end{subfigure}
  \hfill
  \begin{subfigure}{0.3\textwidth}
    \centering
    \includegraphics[width=\linewidth]{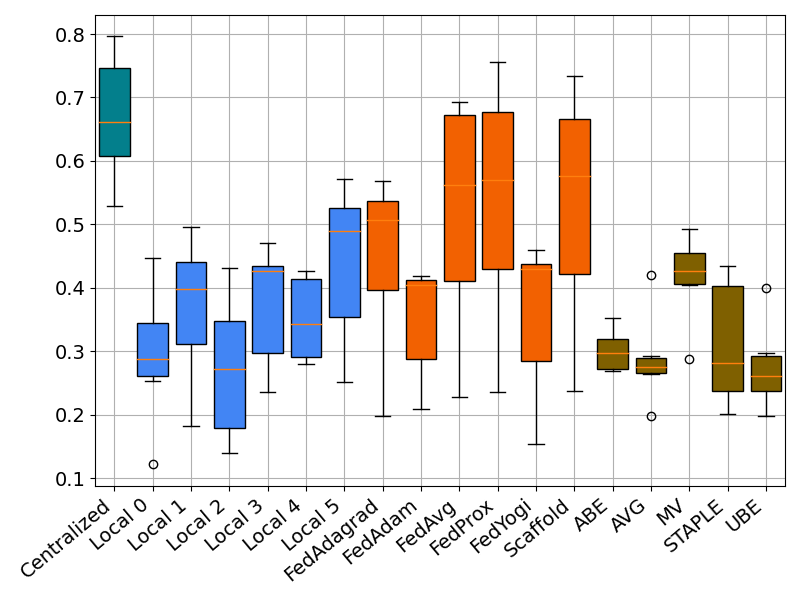}
    \caption{FedKiTS}
    \label{fig:acc_kits}
  \end{subfigure}

  \vspace{12pt}

  \begin{subfigure}{0.3\textwidth}
    \centering
    \includegraphics[width=\linewidth]{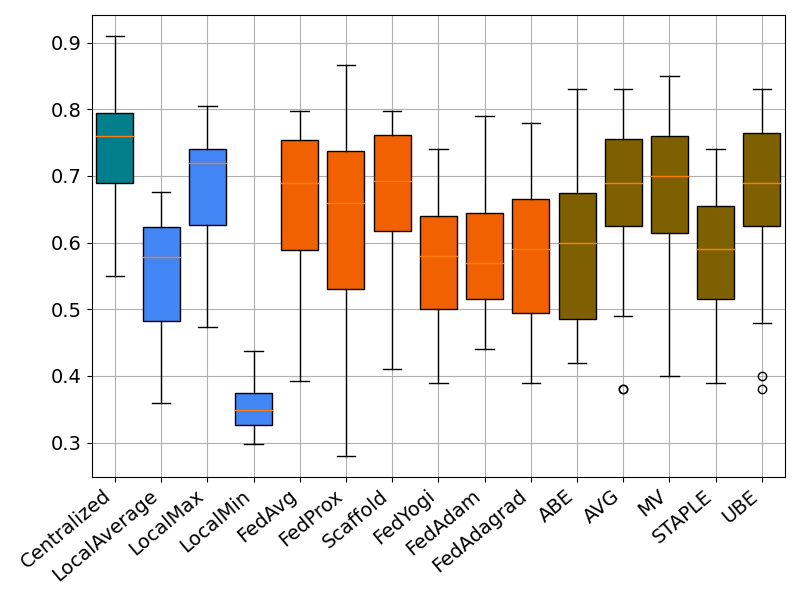}
    \caption{FeTS}
    \label{fig:acc_fets}
  \end{subfigure}

  \caption{Results obtained by centralized learning (\textcolor[HTML]{037f8c}{green}), local learning (\textcolor[HTML]{4285f4}{blue}), federated learning (\textcolor[HTML]{f26101}{orange}), and consensus-based learning (\textcolor[HTML]{7f6000}{brown}) methods. The boxplot represents the accuracy among test sets for centralized learning, local models, and CL methods. For FeTS, local accuracy results are aggregated due to the large number of clients.}
  \label{fig:datasetsaccs}
\end{figure*}

Collaborative approaches yielded a better average performance than locally trained models in five out of seven benchmarks, with the best and worst performance ratios achieved respectively for FedISIC (1.42$\times$ increase) and FedHeart (0.88$\times$ decrease).
On the other hand, collaborative methods had on average poorer performance compared to pooling all data in a centralized training scenario (average decrease of 0.83$\times$). 
This result is in line with the literature on the effectiveness of both FL and CBL methods, known to improve upon individual learners while maintaining comparable performance to pooled datasets.
On average, as shown in Figure~\ref{fig:datasetsaccs}, CBL approaches outperformed FL in four benchmarks (average factor 1.18$\times$), had comparable performance in FeTS (factor 1.03$\times$) and FedISIC (factor 0.97$\times$), and had poorer performance in FedKITS (factor 0.70$\times$). However, tests for the difference in the mean performance across all benchmarks between CBL and FL methods did not achieve statistical significance (Wilcoxon signed-rank test $p$-value=0.22).

We then checked whether there could be differences at the level of task types and of individual benchmarks. CBL outperformed FL in all three task categories (classification, segmentation, survival), however none of these differences were statistically significant (see Table~\ref{tab:tasks}). 

\begin{table*}
\centering
\caption{Summary results for multiple Mann-Whitney U-tests used to assess differences between the performance of consensus-based learning (CBL) and federated learning (FL) for different types of ML tasks. 95\% confidence intervals (CI) for the average performance ratio, obtained via unpaired bootstrap, are reported alongside the corresponding $p$-values. CBL outperforms FL across all tasks, however no difference was statistically significant.}
\begin{tabular}{@{}l c c c c @{}}
\toprule
Task & Average performance ratio $\frac{\text{CBL}}{\text{FL}}$ & 95\% CI & U statistic & $p$-value \\
\midrule
\textsc{Classification} & $1.11$ & $[1.014, 1.228]$ & $18$ & {.26}\\
\textsc{Segmentation} & $1.02$ & $[0.956, 1.099]$ & $17$ & {.79}\\
\textsc{Survival} & $1.06$ & $[1.003, 1.113]$ & $16$ & {.10}\\

\bottomrule
\end{tabular}

\label{tab:tasks}
\end{table*}

Averaging the performance over all client test sets for a given learning method, we statistically assessed whether CBL methods performed worse than FL on average (see Table~\ref{tab:pvalues-kw}). 

\begin{table*}
\centering
\caption{Summary results for multiple Mann-Whitney U-tests used to assess differences between the performance of consensus-based learning (CBL) and federated learning (FL) across benchmarks. 95\% confidence intervals (CI) for the average performance ratio, obtained via unpaired bootstrap, are reported alongside the corresponding $p$-values. The only significant differences are observed in FedIXI, where CBL is associated with better performances than FL, and FedKITS, where CBL is associated with poorer performances than FL. The statistical significance does not survive after Bonferroni correction for multiple comparisons across tasks. }
	\begin{tabular}{@{}l S[table-format=1.2] c S[table-format=1.2] S[table-format=1.2] @{}}
\toprule
Experiment & {Average performance ratio ($\frac{\text{CBL}}{\text{FL}}$)} & {95\% CI} & {U statistic} & {$p$-value} \\
\midrule
FedProstate & {$1.12\times$} & {[1.015, 1.245]} & ${22.5}$ & {.20}\\
FedHeart & {$1.36\times$} & {[1.070, 1.801]} & ${19}$ & {.16}\\
FedIXI & {$1.18\times$} & {[1.105, 1.258]} & ${29}$ & {.01}\\
FedISIC & {$0.97\times$} & {[0.950, 1.001]} & ${5}$ & {.17}\\
FedTCGA-BRCA & {$1.06\times$} & {[1.003, 1.112]} & ${10}$ & {.10}\\
FedKITS & {$0.70\times$} & {[0.581, 0.855]} & ${2}$ & {.02}\\
FeTS & {$1.03\times$} & {[0.957, 1.107]} & ${19}$ & {.54}\\
\bottomrule
\end{tabular}

\label{tab:pvalues-kw}
\end{table*}

The only two statistically significant outcomes were in the IXI and KITS benchmarks: CBL had better performance for IXI (factor 1.18$\times$, Mann-Whitney $p$-value=0.009), but poorer performance for KITS (factor 0.70$\times$, Mann-Whitney $p$-value=0.02). It is important to note that neither results maintain statistical significance after Bonferroni correction. 

The ensemble of these findings shows that, overall, no single collaborative learning paradigm consistently outperforms the others on all benchmarks. 
For example, as already pointed out in the literature, FL performs worse than local learning in the FedIXI benchmark, whereas CBL achieves higher accuracy~\cite{terrail2022flamby}. 
In contrast, for FedKITS, FL methods had better performance compared to local models, while CBL methods had a poorer one.

Furthermore, we identified the best-performing CBL and FL method separately for each benchmark and compared their performance. The best CBL method had a better performance than the best FL method in the FedHeart (factor 1.09$\times$) and FedIXI (factor 1.06$\times$) benchmarks, they had comparable performances in four benchmarks, and the best CBL method performed worse than the best FL method only in the FedKITS benchmark (factor 0.77$\times$). When instead comparing the basic methods for each CL category, i.e. FedAVG for FL and AVG for CBL, we found that CBL had better performance than FedAVG in three out of seven benchmarks (average factor 1.37$\times$), had comparable performance in the FeTS (factor 1.01$\times$) and FedISIC (factor 1.00$\times$) benchmarks, and had worse performance in the FedProstate (factor 0.94$\times$) and FedKITS (factor 0.55$\times$) benchmarks. The average performance across all client test sets, for both commonly-used and best-performing methods, is summarized in Table~\ref{tab:perf}.

\begin{table*}
\centering
\caption{Summary performance metrics for commonly-used (FedAVG and AVG) and best-performing training methods within the CBL and FL groups, computed separately for each benchmark. 95\% confidence intervals (CI) for each metric, obtained via unpaired bootstrap, are reported in brackets. Neither of the two paradigms yields better performance in a consistent and statistically significant way. }
\resizebox{\textwidth}{!}{
\begin{tabular}{l l l c c l c c}
\toprule
Experiment & Paradigm & \makecell[l]{Best \\ method} & Metric & 95\% CI & \makecell[l]{Common\\ method} & Metric & 95\% CI\\
\midrule
\multirow{2}{*}{FedProstate}    &  CBL  &  UBE          &  $0.81$   &  $[0.758, 0.860]$ &  AVG     &  $0.75$ &  $[0.680, 0.817]$ \\
                                &  FL   &  Scaffold     &  $0.82$   &  $[0.760, 0.868]$ &  FedAVG  &  $0.80$ &  $[0.743, 0.850]$ \\
\multirow{2}{*}{FedHeart}       &  CBL  &  UBE          &  $0.78$   &  $[0.684, 0.924]$ &  AVG     &  $0.64$ &  $[0.416, 0.869]$ \\
                                &  FL   &  FedAdam      &  $0.72$   &  $[0.667, 0.769]$ &  FedAVG  &  $0.36$ &  $[0.132, 0.574]$ \\
\multirow{2}{*}{FedIXI}         &  CBL  &  MV           &  $0.96$   &  $[0.959, 0.964]$ &  AVG     &  $0.90$ &  $[0.891, 0.911]$ \\
                                &  FL   &  FedAdagrad   &  $0.90$   &  $[0.896, 0.909]$ &  FedAVG  &  $0.75$ &  $[0.742, 0.751]$ \\
\multirow{2}{*}{FedISIC}        &  CBL  &  AVG          &  $0.62$   &  $[0.522, 0.729]$ &  AVG     &  $0.62$ &  $[0.525, 0.728]$ \\
                                &  FL   &  FedYogi      &  $0.65$   &  $[0.518, 0.761]$ &  FedAVG  &  $0.62$ &  $[0.512, 0.726]$ \\
\multirow{2}{*}{FedTCGA-BRCA}   &  CBL  &  AVG          &  $0.79$   &  $[0.643, 0.927]$ &  AVG     &  $0.79$ &  $[0.633, 0.927]$ \\
                                &  FL   &  FedAdagrad   &  $0.78$   &  $[0.619, 0.921]$ &  FedAVG  &  $0.70$ &  $[0.553, 0.846]$ \\
\multirow{2}{*}{FedKITS}        &  CBL  &  MV           &  $0.42$   &  $[0.362, 0.461]$ &  AVG     &  $0.29$ &  $[0.239, 0.348]$ \\
                                &  FL   &  FedProx      &  $0.54$   &  $[0.384, 0.672]$ &  FedAVG  &  $0.52$ &  $[0.375, 0.648]$ \\
\multirow{2}{*}{FeTS}           &  CBL  &  MV           &  $0.67$   &  $[0.617, 0.715]$ &  AVG     &  $0.66$ &  $[0.612, 0.711]$ \\
                                &  FL   &  Scaffold     &  $0.67$   &  $[0.630, 0.715]$ &  FedAVG  &  $0.66$ &  $[0.609, 0.701]$ \\
\bottomrule
\end{tabular}
}

\label{tab:perf}
\end{table*}

Having assessed that no paradigm outperforms the other in terms of task-specific performance metrics, we now look at their cost-effectiveness. 
The numerical evaluation of the cost-effectiveness and sustainability of CBL and FL, presented in Table~\ref{tab:costeffect}, shows that the training of CBL methods is 8-30 times faster than the one required by the federated methods (avg. 15$\times$ decrease, P=.02 Wilcoxon Signed-Rank Test). 
For the same reason, the bandwidth usage of CBL is 35 to 345 times lower than that of FL (avg. 131$\times$ decrease, P=.02 Wilcoxon Signed-Rank Test).

\begin{table*}[ht]
    \centering
    \caption{Comparison of training time and bandwidth usage between federated learning (FL) and consensus-based learning (CBL). For both metrics, CBL is far less costly than FL. 95\% confidence intervals (CI), obtained via unpaired bootstrap, are reported in brackets for the training time ratio. No CI is reported for the bandwidth ratio, as bandwidth usage is a deterministic quantity given fixed hardware and network conditions, and therefore takes a single value across runs.}
    \resizebox{\textwidth}{!}{
    \begin{NiceTabular}{@{} r||ccc|c|ccc @{}}
        \toprule
        & \multicolumn{4}{c|}{\textbf{Training Time} [min]}
        & \multicolumn{3}{c}{\textbf{Bandwidth} [MB]} \\
        \midrule
        & FL & CBL & \multicolumn{1}{c}{FL Slow factor} & \multicolumn{1}{c|}{95\% CI} & FL & CBL & \multicolumn{1}{c}{FL Increment} \\
        \midrule
        FedProstate & $1.2\times10^3$ & $1.1 \times10^2$ &  $8\times$ & $[4.6, 15.6]$ & $4.0 \times 10^3$  & $1.15 \times 10^2$ & $35\times$ \\

        FedHeart& $0.15 $& $0.017$ &  $9\times$ & $[5.8, 15.7]$ & $13.1$ & $0.09$ &  $146\times$ \\

        FedIXI & $5.5$ &  $0.65$  &  $8\times$ & $[5.2, 22.0]$ & $4.5 \times 10^2$  & $3$ & $150\times$ \\

        FedISIC & $193$ &  $10.2$ & $19\times$ & $[10.0, 50.3]$ & $4.06 \times 10^4$  & $1.18 \times 10^2$ & $345\times$ \\

        FedTCGA-BRCA & $0.455$ & $0.017$ & $26\times$ & $[21.2, 34.8]$ & $5.42$ & $0.15$ & $36\times$ \\

        FedKITS & $1581$ &  $54.1$ & $29\times$ & $[21.9, 37.2]$ & $8.8 \times 10^4$  & $7.4 \times 10^{2}$ & $120\times$ \\

        FeTS & $3.4 \times10^{4} $ &  $2.9 \times 10^{3}$ & $12\times$ & $[10.0, 15.1]$ & $3.7 \times 10^4$  &  $4.5 \times 10^2$ & $83\times$ \\
        \bottomrule
    \end{NiceTabular}
    }

    \label{tab:costeffect}
\end{table*}

\section*{Discussion}

Our results show that for typical medical data analysis tasks, CBL leads to model performance on par with FL, while requiring a fraction of computational costs and bandwidth usage. 
Beyond the immediate cost savings, these reductions carry broader sustainability implications: lower computational demand translates into reduced energy consumption and associated ecological footprint, while lower infrastructure and maintenance costs improve the economic accessibility of collaborative AI for healthcare institutions operating with limited resources, including those in low- and middle-income settings. 

The key advantage of CBL is its asynchronous training approach, which simplifies collaboration among partners since hospitals simply need to train their models locally to join the collaboration. Moreover, the withdrawal (or introduction) of a participant can be simply achieved by removing (or adding) their local model without the need for complex procedures or re-training. The complexity of local models can also be adapted to the availability of local resources, thus mitigating the problem of hardware heterogeneity in federated setups, and promoting the adoption of AI in healthcare. In FL, besides tuning the hyperparameters of the local models, it is also often necessary to tune method-specific hyperparameters, which increases the task complexity compared to CBL. Using the same hyperparameters on all FL methods can lead to variability in the accuracy across methods. In contrast, CBL methods exhibit lower variance among results.

Data heterogeneity, a common issue in healthcare data, may also play a role in the choice of CBL over FL approaches. Indeed, while the generalization properties of CBL may improve when ensembling models trained on heterogeneous datasets, it is known that the distributed optimization procedure of FL suffers from data heterogeneity, which causes degradation in convergence and accuracy. Such difference suggests that adopting CBL could be beneficial in collaborations among hospitals with heterogeneous data, reflecting for example varying geographical location or image acquisition techniques. Nonetheless, when local datasets are excessively diverse or unbalanced, the disagreement among local models can lower the confidence of the CBL consensus, e.g. resulting in a less decisive majority vote or a higher-entropy averaged prediction, which can in turn be used as a signal to flag collaborations affected by substantial data heterogeneity.

The analysis addressed in this paper focuses on the cost-effectiveness of CBL to boost the deployment of AI in real-life healthcare applications. 
However, CBL does not necessarily provide quantifiable privacy guarantees. 
While cryptographic approaches such as secure aggregation and multi-party computation are widely investigated in FL and are now implemented in most software, the analysis of CBL from a privacy-preserving perspective is less explored, with some studies showing that privacy-preserving techniques can be adopted for CBL, with a potentially smaller impact on the final model accuracy when compared to FL~\cite{chaudhari2023safenet}. 
A finer-grained comparison reveals that the two paradigms interact differently with specific privacy-enhancing techniques, each entailing its own trade-offs. For differential privacy (DP), both FL and CBL admit a local and a central (global) level of protection, but with different implications. In FL, DP affects the entire training process, since noise must be injected at every communication round, coupling the achievable privacy budget to the number of rounds and thus to convergence and accuracy. In CBL, the local level coincides with standard DP training of a stand-alone model, while the central level can instead be applied a posteriori to the consensus step at inference time, leaving local training unaffected. Regarding secure aggregation and encryption, in FL the aggregated global model is disclosed to all participants at every round, potentially exposing it to model-inversion or membership-inference attacks unless further cryptographic protection is applied; in CBL, local models are never merged into a single shared model, so preserving the confidentiality of the consensus instead requires performing inference in encrypted form, which may add further computational complexity to the workflow. Finally, machine unlearning highlights a further distinction between the two paradigms: because client contributions are entangled across FL's iterative rounds, removing one participant's influence generally requires re-involving the entire federation, whereas CBL allows a center's model to be withdrawn from the consensus and retrained independently, offering a more direct and self-contained route to compliance with data removal requirements.

Although CBL appears as a promising and cost-effective paradigm across the wide range of benchmarks considered in this study, it has not yet been extensively validated in complex, real-world multi-institutional deployments, and unforeseen challenges may emerge as this paradigm is adopted at scale.

With this work, we hope to raise awareness of the importance of cost-effective collaborative learning paradigms in the real-world deployment of AI models on medical tasks. Starting from the empirical results presented here, additional work should be devoted to developing quantitative measures of a collaborative system’s effectiveness before deploying its infrastructure. Such a theory would allow the identification of optimal collaborative paradigms based on accounting for several aspects, such as client availability, data heterogeneity, and security requirements. Moreover, in evaluating the cost-effectiveness, estimation of the CO$_2$ emissions and energy cost of the training would bring added value to the analysis.

\section*{Methods}



\subsection*{Benchmark dataset and tasks}

All the data used in our benchmark are publicly available and anonymized, except for a small private dataset of 36 MRIs from cancer patients undergoing active surveillance used as external validation for the FedProstate benchmark, for which we obtained the necessary IRB approval (REC 23/NW/0105: Guy’s Cancer Cohort ethics) and for which informed consent was obtained from all subjects and/or their legal guardian(s). All methods were carried out in accordance with relevant guidelines and regulations.
The training data for the FedProstate benchmark were obtained by aggregating three publicly available datasets from the Decathlon, Promise12, and ProstateX challenges~\cite{antonelli2022medical,litjens2014evaluation,armato2018prostatex}. The external validation set for this task uses a private, retrospectively-collected dataset of patients undergoing active surveillance as explained above. The FeTS dataset is taken from the homonymous tumor segmentation challenge~\cite{pati2022federated}, while the other datasets are obtained from the FLamby benchmark~\cite{terrail2022flamby}. A summary of the tasks and datasets is given in Table~\ref{tab:dataset}

\subsection*{Collaborative Learning Methods}\label{subsec:methods}
We consider a collaborative setting with ${M}$ clients. A dataset $\mathcal{D}$ belonging to client $i$ is composed of data samples $\mathcal{D}_i = \{z_{k,i}\}_{k=1}^{N_i}$, being $N_i$ the dataset size.
We consider a model $f$ with parameters $\theta$, a loss function $\mathcal{L}$, and we denote the prediction of a data instance $z$ by $h = f(z, \theta)$. Table~\ref{tab:methods} presents a summary and main references of both FL and CBL methodologies benchmarked in this paper.

\begin{table*}
\centering
\caption{Collaborative learning (CL) methods evaluated in the benchmark: six methods for federated learning (FL) and five methods for consensus-based learning (CBL). }
\begin{tabular}{@{}l l l @{}}
\toprule
Method & CL Paradigm & Reference \\
\midrule
\textsc{FedAvg} & FL & McMahan et. Al., 2017 \cite{FedAvg}\\
\textsc{FedProx} & FL & Li et Al., 2018 \cite{FedProx}\\
\textsc{scaffold} & FL & Karimireddy et Al., 2020 \cite{karimireddy2020scaffold}\\
\textsc{FedAdam} & FL & Reddi et Al., 2020 \cite{reddi2020adaptive}\\
\textsc{FedAdagrad} & FL & Reddi et Al., 2020 \cite{reddi2020adaptive}\\
\textsc{FedYogi} & FL & Reddi et Al., 2020 \cite{reddi2020adaptive}\\
\textsc{Avg} & CBL & Guha et Al., 2019 \cite{guha2019one}\\
\textsc{Mv} & CBL & Safdar et Al., 2021 \cite{safdar2021majority}\\
\textsc{Staple} & CBL & Warfield et Al., 2004 \cite{warfield2004simultaneous} \\
\textsc{Ube} & CBL & \textit{Inspired by} Ruta et Gabrys, 2000 \cite{ruta2000overview} \\
\textsc{Abe} & CBL & \textit{Inspired by} Ruta et Gabrys, 2000 \cite{ruta2000overview} \\
\bottomrule
\end{tabular}

\label{tab:methods}
\end{table*}

\subsubsection*{Federated Learning}
\textbf{FL} is a collaborative paradigm associated with the optimization of a loss distributed among $M$ clients defined as $\mathcal{L}(\theta) = \sum_{i=1}^{M} p_i\mathcal{L}_i(\theta_i)$, where
the losses of local models ($\theta_i$) are averaged by using the weights $p_i$. Solving this optimization problem leads to global model parameters $\theta_g$. We consider here a comprehensive panel of state-of-the-art optimization approaches, which are at the core of the FL literature:
\begin{itemize}
     \item \textbf{\textsc{FedAvg}}\cite{FedAvg} is the backbone of FL optimization, and is based on an iterative process where, at each optimization round $r$,  clients execute a fixed number of local stochastic gradient descent steps and send the partially optimized model $\theta_i^r$ to the server. The server averages the received models according to the weights $p_i$ to obtain a global one, $\theta_g^{r+1}$. 
    The global model is then sent to the clients to initialize the next optimization round. 
    \item \textbf{\textsc{FedProx}}\cite{FedProx} tackles the problem of federated optimization with data heterogeneity across clients. This approach extends \textsc{FedAvg} by introducing a proximal term to the local objective function to penalize model drift from the global optimization during local training. The proximal term is controlled by a trade-off hyperparameter. 
    \item \textbf{{\textsc{SCAFFOLD}}\cite{karimireddy2020scaffold}} addresses the limitations of \textsc{FedAvg} in scenarios with heterogeneous data by utilizing control variates, which effectively reduces variance and corrects for client-drift in the local updates. To achieve this, \textsc{SCAFFOLD} maintains a state for each client (client control variate) and the server (server control variate).
    \item \textbf{\textsc{FedAdam}, \textsc{FedYogi}, \textsc{FedAdagrad}\cite{reddi2020adaptive}} are adaptations of Adam
    , Yogi
    , and Adagrad
    optimizers, designed to suit the federated optimization setup. 
\end{itemize}

\subsubsection*{Consensus-Based Learning}
\textbf{CBL}, on the other hand, is a class of machine learning algorithms relying on the concept of \emph{ensembling}, a widely-explored approach that 
consists of obtaining robust predictions by aggregating decisions obtained by independently trained weak predictors. 

The CBL paradigm is based on a \emph{training} phase in which $M$ models $f(\cdot, \theta_i)$ are independently trained on separated data collections $\mathcal{D}_i$, by minimizing local objective functions $\mathcal{L}_i$. 
These models are subsequently collected and, for a given test data $z'$ at \emph{inference} time, the predictions $h_i(z') = f(z', \theta_i)$ from all the clients' models are computed and aggregated by applying an ensembling (or fusion) strategy:
\begin{equation}\label{eq:ensembling}
     h_{z'} = \texttt{ensembling}(\{h_i(z')\}|_{i=1}^M). 
\end{equation}

Typical ensembling methods proposed in the literature are: 

\textbf{Majority voting} (\textsc{Mv}) \cite{safdar2021majority} is often used in classification tasks, aggregates predictions by selecting the most commonly predicted class among the experts.

\textbf{\textsc{Staple}} \cite{warfield2004simultaneous} is an algorithm based on expectation-maximization which, iteratively, first computes a weighted average of each local prediction, and then assigns a performance level to each client's segmentation, which will be used as weights for the next step.

\textbf{Decision averaging} (DA) \cite{adiga2021all} is an approach based on probabilistic principles: given different datasets $\mathcal{D}_1,\dots\mathcal{D}_M$ and associated local models  $f(\cdot, \theta_1),\dots f(\cdot, \theta_M)$, ensembling is obtained by weighing each local model's prediction by the probability of observing the input datapoint in the local model's data distribution $p(z'\in \mathcal{D}_i)$. 
Different CBL algorithms can be obtained based on the estimation of the probability $p(z'\in \mathcal{D}_i)$.
In this work, among the possible DA algorithms, we consider the following:
\begin{itemize}
    \item \textbf{Average} (\textsc{Avg})  approximates $p(z'\in \mathcal{D}_i)$ as a uniform distribution. This strategy can be adopted in all tasks, and for classification and segmentation problems we consider the distribution probability obtained through a \textit{softmax} function.
    \item \textbf{Uncertainty based ensembling} (\textsc{Ube}) approximates the probability $p(z'\in \mathcal{D}_i)$ as the uncertainty of the local model on the prediction task. \textsc{Ube} defines averaging weights based on the model uncertainty quantified by the total element-wise variance at inference time.
    \item \textbf{Autoencoder based ensembling} (\textsc{Abe}) computes a proxy for the probability $p(z'\in \mathcal{D}_i)$ by modeling the variability of the local dataset through autoencoders trained by each client on the respective data. At inference time, weights are defined as the reconstruction error $e_i$ on the testing data point $z$.
\end{itemize}

\subsection*{Benchmark design}

We split each dataset in the benchmark into training and testing partitions. For FeTS and FedProstate, we used respectively a 4-fold and a 5-fold cross-validation while for the other datasets we ran all the experiments three times with different seeds following previous work~\cite{terrail2022flamby}. All model parameters were randomly initialized. The obtained results are the average among the runs. The number of folds/seeds was chosen to balance the computational cost and runtime of the overall benchmark, which spans multiple datasets, tasks and methods, against the need to adequately characterize the statistical variability of the results.

For FL, a federated infrastructure was simulated using the software FedBioMed version 4.1. For each FL method, the final global model was collected along with the local models independently trained by each client. The local models were used to estimate the local model performance and for generating the predictions subsequently aggregated with the CBL methods. As upper-bound for the comparison, a centralized model was trained by pooling together all the local training sets. All FL and CBL experiments for a given benchmark, including the training time and bandwidth measurements reported in Table~\ref{tab:costeffect}, were executed under identical hardware and (simulated) network conditions to ensure a fair comparison between the two paradigms.

To ensure fairness of the comparison, for each dataset we calibrated the total number of training steps to be executed in each experiment. Specifically, a number $E$ of epochs was defined a priori, corresponding to the number of training epochs executed by the centralized model. For the local training, each client executes $\frac{E}{M}$ epochs, being $M$ the number of clients in that configuration. For the federated strategies, we fine-tuned the number of local stochastic gradient descent steps $s$ executed at each round, while the number of rounds was defined as follows: 
\begin{equation}
R = \frac{E\cdot N_{train}}{M\cdot B \cdot s}, 
\end{equation}
where $B$ is the batch size and $N_{train}$ is the total number of samples in the training set.

\subsection*{Statistical analysis}
We used a Wilcoxon signed-rank test to assess the difference in the average performance of all FL compared to the average of all CBL methods on a benchmark-by-benchmark basis. To compare the performance of FL against CBL methods on benchmark-by-benchmark basis, but without averaging over sub-methods, we used one Mann-Whitney U-test per benchmark type, i.e. seven in total, with Bonferroni correction to account for the multiple tests. To compare performance aggregated at the level of task categories, we first averaged the performance of FL and CBL over their respective sub-methods, then we computed the performance ratio $\frac{CBL}{FL}$ and used one Mann-Whitney U-test per task category (three in total), with Bonferroni correction. For each benchmark and each task category, 95\% confidence intervals for the average performance ratio $\frac{CBL}{FL}$ (Tables~\ref{tab:tasks} and~\ref{tab:pvalues-kw}) were obtained via unpaired bootstrap. Similarly, 95\% confidence intervals for the performance metrics of the commonly-used and best-performing methods (Table~\ref{tab:perf}) were obtained via unpaired bootstrap. Finally, to compare the wallclock runtime and bandwidth usage of FL and CBL methods on a benchmark-by-benchmark basis we used a Wilcoxon signed-rank test. A 95\% confidence interval for the training time ratio (Table~\ref{tab:costeffect}) was obtained via unpaired bootstrap; no such interval is reported for the bandwidth ratio, since bandwidth usage is deterministic given fixed hardware and network conditions.

\bibliography{sn-bibliography}

@article{caldas2018leaf,
  title={Leaf: A benchmark for federated settings},
  author={Caldas, Sebastian and Duddu, Sai Meher Karthik and Wu, Peter and Li, Tian and Kone{\v{c}}n{\`y}, Jakub and McMahan, H Brendan and Smith, Virginia and Talwalkar, Ameet},
  journal={arXiv preprint arXiv:1812.01097},
  year={2018}
}

@InProceedings{FedAvg, 
    title = {{Communication-Efficient Learning of Deep Networks from Decentralized Data}}, 
    author = {Brendan McMahan and Eider Moore and Daniel Ramage and Seth Hampson and Blaise Aguera y Arcas}, 
    booktitle = {ICML 2017},  
    year = {2017},     
}

@article{FedProx,
	archivePrefix = {arXiv},
	arxivId = {1812.06127},
	author = {Li, Tian and Sahu, Anit Kumar and Zaheer, Manzil and Sanjabi, Maziar and Talwalkar, Ameet and Smith, Virginia},
	eprint = {1812.06127},
	journal = {Proceedings of the 1 st Adaptive \& Multitask Learning Workshop, Long Beach, California, 2019},
	pages = {1--28},
	title = {{Federated Optimization in Heterogeneous Networks}},
	year = {2018}
}

@inproceedings{karimireddy2020scaffold,
  title={Scaffold: Stochastic controlled averaging for federated learning},
  author={Karimireddy, Sai Praneeth and Kale, Satyen and Mohri, Mehryar and Reddi, Sashank and Stich, Sebastian and Suresh, Ananda Theertha},
  booktitle={International conference on machine learning},
  pages={5132--5143},
  year={2020},
  organization={PMLR}
}

@inproceedings{safdar2021majority,
  title={A majority voting based ensemble approach of deep learning classifiers for automated melanoma detection},
  author={Safdar, Khadija and Akbar, Shahzad and Shoukat, Ayesha},
  booktitle={2021 International Conference on Innovative Computing (ICIC)},
  pages={1--6},
  year={2021},
  organization={IEEE}
}

@inproceedings{adiga2021all,
  title={All models are useful: Bayesian ensembling for robust high resolution covid-19 forecasting},
  author={Adiga, Aniruddha and Wang, Lijing and Hurt, Benjamin and Peddireddy, Akhil and Porebski, Przemyslaw and Venkatramanan, Srinivasan and Lewis, Bryan Leroy and Marathe, Madhav},
  booktitle={Proceedings of the 27th ACM SIGKDD Conference on Knowledge Discovery \& Data Mining},
  pages={2505--2513},
  year={2021}
}

@article{warfield2004simultaneous,
  title={Simultaneous truth and performance level estimation (STAPLE): an algorithm for the validation of image segmentation},
  author={Warfield, Simon K and Zou, Kelly H and Wells, William M},
  journal={IEEE transactions on medical imaging},
  volume={23},
  number={7},
  pages={903--921},
  year={2004},
  publisher={IEEE}
}

@article{lee2020federated,
  title={Federated learning on clinical benchmark data: performance assessment},
  author={Lee, Geun Hyeong and Shin, Soo-Yong},
  journal={Journal of medical Internet research},
  volume={22},
  number={10},
  pages={e20891},
  year={2020},
  publisher={JMIR Publications Toronto, Canada}
}

@article{chen2022pfl,
  title={pFL-bench: A comprehensive benchmark for personalized federated learning},
  author={Chen, Daoyuan and Gao, Dawei and Kuang, Weirui and Li, Yaliang and Ding, Bolin},
  journal={Advances in Neural Information Processing Systems},
  volume={35},
  pages={9344--9360},
  year={2022}
}

@article{terrail2022flamby,
  title={Flamby: Datasets and benchmarks for cross-silo federated learning in realistic healthcare settings},
  author={du Terrail, Jean Ogier du and Ayed, Samy-Safwan and Cyffers, Edwige and Grimberg, Felix and He, Chaoyang and Loeb, Regis and Mangold, Paul and Marchand, Tanguy and Marfoq, Othmane and Mushtaq, Erum and others},
  journal={arXiv preprint arXiv:2210.04620},
  year={2022}
}

@article{kairouz2021advances,
  title={Advances and open problems in federated learning},
  author={Kairouz, Peter and McMahan, H Brendan and Avent, Brendan and Bellet, Aur{\'e}lien and Bennis, Mehdi and Bhagoji, Arjun Nitin and Bonawitz, Kallista and Charles, Zachary and Cormode, Graham and Cummings, Rachel and others},
  journal={Foundations and Trends (R) in Machine Learning},
  volume={14},
  number={1--2},
  pages={1--210},
  year={2021},
  publisher={Now Publishers, Inc.}
}

@article{reddi2020adaptive,
  title={Adaptive federated optimization},
  author={Reddi, Sashank and Charles, Zachary and Zaheer, Manzil and Garrett, Zachary and Rush, Keith and Kone{\v{c}}n{\`y}, Jakub and Kumar, Sanjiv and McMahan, H Brendan},
  journal={arXiv preprint arXiv:2003.00295},
  year={2020}
}

@article{wang2020federated,
  title={Federated learning with matched averaging},
  author={Wang, Hongyi and Yurochkin, Mikhail and Sun, Yuekai and Papailiopoulos, Dimitris and Khazaeni, Yasaman},
  journal={arXiv preprint arXiv:2002.06440},
  year={2020}
}

@article{joshi2022federated,
  title={Federated learning for healthcare domain-pipeline, applications and challenges},
  author={Joshi, Madhura and Pal, Ankit and Sankarasubbu, Malaikannan},
  journal={ACM Transactions on Computing for Healthcare},
  volume={3},
  number={4},
  pages={1--36},
  year={2022},
  publisher={ACM New York, NY}
}

@article{antonelli2022medical,
  title={The medical segmentation decathlon},
  author={Antonelli, Michela and Reinke, Annika and Bakas, Spyridon and Farahani, Keyvan and Kopp-Schneider, Annette and Landman, Bennett A and Litjens, Geert and Menze, Bjoern and Ronneberger, Olaf and Summers, Ronald M and others},
  journal={Nature communications},
  volume={13},
  number={1},
  pages={4128},
  year={2022},
  publisher={Nature Publishing Group UK London}
}

@article{litjens2014evaluation,
  title={Evaluation of prostate segmentation algorithms for MRI: the PROMISE12 challenge},
  author={Litjens, Geert and Toth, Robert and Van De Ven, Wendy and Hoeks, Caroline and Kerkstra, Sjoerd and Van Ginneken, Bram and Vincent, Graham and Guillard, Gwenael and Birbeck, Neil and Zhang, Jindang and others},
  journal={Medical image analysis},
  volume={18},
  number={2},
  pages={359--373},
  year={2014},
  publisher={Elsevier}
}

@article{armato2018prostatex,
  title={PROSTATEx Challenges for computerized classification of prostate lesions from multiparametric magnetic resonance images},
  author={Armato III, Samuel G and Huisman, Henkjan and Drukker, Karen and Hadjiiski, Lubomir and Kirby, Justin S and Petrick, Nicholas and Redmond, George and Giger, Maryellen L and Cha, Kenny and Mamonov, Artem and others},
  journal={Journal of Medical Imaging},
  volume={5},
  number={4},
  pages={044501--044501},
  year={2018},
  publisher={Society of Photo-Optical Instrumentation Engineers}
}

@article{pati2021federated,
  title={The federated tumor segmentation (fets) challenge},
  author={Pati, Sarthak and Baid, Ujjwal and Zenk, Maximilian and Edwards, Brandon and Sheller, Micah and Reina, G Anthony and Foley, Patrick and Gruzdev, Alexey and Martin, Jason and Albarqouni, Shadi and others},
  journal={arXiv preprint arXiv:2105.05874},
  year={2021}
}

@article{guha2019one,
  title={One-shot federated learning},
  author={Guha, Neel and Talwalkar, Ameet and Smith, Virginia},
  journal={arXiv preprint arXiv:1902.11175},
  year={2019}
}

@article{rieke2020future,
  title={The future of digital health with federated learning},
  author={Rieke, Nicola and Hancox, Jonny and Li, Wenqi and Milletari, Fausto and Roth, Holger R and Albarqouni, Shadi and Bakas, Spyridon and Galtier, Mathieu N and Landman, Bennett A and Maier-Hein, Klaus and others},
  journal={NPJ digital medicine},
  volume={3},
  number={1},
  pages={119},
  year={2020},
  publisher={Nature Publishing Group UK London}
}

@article{innocenti2023benchmarking,
  title={Benchmarking Collaborative Learning Methods Cost-Effectiveness for Prostate Segmentation},
  author={Innocenti, Lucia and Antonelli, Michela and Cremonesi, Francesco and Sarhan, Kenaan and Granados, Alejandro and Goh, Vicky and Ourselin, Sebastien and Lorenzi, Marco},
  journal={arXiv preprint arXiv:2309.17097},
  year={2023}
}

@inproceedings{chaudhari2023safenet,
  title={SafeNet: The Unreasonable Effectiveness of Ensembles in Private Collaborative Learning},
  author={Chaudhari, Harsh and Jagielski, Matthew and Oprea, Alina},
  booktitle={2023 IEEE Conference on Secure and Trustworthy Machine Learning (SaTML)},
  pages={176--196},
  year={2023},
  organization={IEEE}
}

@article{nature2021coll,
  title={Collaborative learning without sharing data},
  journal={Nature Machine Intelligence},
  volume={3},
  number={459},
  year={2021},
  note    = {Editorial},
  author = {Editorial Board}
}

@article{gupta2023collaborative,
  title={Collaborative privacy-preserving approaches for distributed deep learning using multi-institutional data},
  author={Gupta, Sharut and Kumar, Sourav and Chang, Ken and Lu, Charles and Singh, Praveer and Kalpathy-Cramer, Jayashree},
  journal={RadioGraphics},
  volume={43},
  number={4},
  pages={e220107},
  year={2023},
  publisher={Radiological Society of North America}
}

@article{ruta2000overview,
  title={An overview of classifier fusion methods},
  author={Ruta, Dymitr and Gabrys, Bogdan},
  journal={Computing and Information systems},
  volume={7},
  number={1},
  pages={1--10},
  year={2000}
}

@article{sheller2020federated,
  title={Federated learning in medicine: facilitating multi-institutional collaborations without sharing patient data},
  author={Sheller, Micah J and Edwards, Brandon and Reina, G Anthony and Martin, Jason and Pati, Sarthak and Kotrotsou, Aikaterini and Milchenko, Mikhail and Xu, Weilin and Marcus, Daniel and Colen, Rivka R and others},
  journal={Scientific reports},
  volume={10},
  number={1},
  pages={12598},
  year={2020},
  publisher={Nature Publishing Group UK London}
}

@article{pati2022federated,
  title={Federated learning enables big data for rare cancer boundary detection},
  author={Pati, Sarthak and Baid, Ujjwal and Edwards, Brandon and Sheller, Micah and Wang, Shih-Han and Reina, G Anthony and Foley, Patrick and Gruzdev, Alexey and Karkada, Deepthi and Davatzikos, Christos and others},
  journal={Nature communications},
  volume={13},
  number={1},
  pages={7346},
  year={2022},
  publisher={Nature Publishing Group UK London}
}

@article{usynin2021adversarial,
  title={Adversarial interference and its mitigations in privacy-preserving collaborative machine learning},
  author={Usynin, Dmitrii and Ziller, Alexander and Makowski, Marcus and Braren, Rickmer and Rueckert, Daniel and Glocker, Ben and Kaissis, Georgios and Passerat-Palmbach, Jonathan},
  journal={Nature Machine Intelligence},
  volume={3},
  number={9},
  pages={749--758},
  year={2021},
  publisher={Nature Publishing Group UK London}
}

\section*{Funding declaration}

LI and ML are supported by the 3IA C\^ote d’Azur Investments
in the Future project managed by the National Research Agency (ref.n ANR-15-IDEX-01 and ANR-19-P3IA-0002). ML is funded by the grants (TRAIN  ANR-22-FAI1-0003) and (Fed-BioMed ANR-19-CE45-0006). The other authors have no funding to report. 

\section*{Data Availability}
The data for the FedHeart, FedIXI, FedISIC, FedTCGA-BRCA, and FedKITS benchmarks are publicly available from the FLamby repository~\cite{terrail2022flamby}.
The data for the FeTS benchmark are publicly available from the repository provided in the publication~\cite{pati2021federated}.
The data for the FedProstate benchmark are publicly available from the \emph{medical segmentation decathlon}~\cite{antonelli2022medical}, the \emph{promise12} challenge~\cite{litjens2014evaluation}, and the \emph{prostatex} challenge~\cite{armato2018prostatex}, except for a private external validation dataset. This dataset is available from the corresponding author on reasonable request.

\section*{Author contributions statement}

\begin{itemize}
    \item Conceptualization: LI, SO, VG, MA and ML
    \item Methodology: LI, FC, MA, ML
    \item Investigation: LI, FC, MA, ML
    \item Visualization: LI
    \item Funding acquisition: SO, ML
    \item Supervision: SO, MA, ML
    \item Writing – original draft: LI, FC
    \item Writing – review \& editing: FC, SO, VG, MA and ML
\end{itemize}







\end{document}